\documentclass{bmvc2k}

\usepackage{verbatim}
\usepackage{amsfonts}
\usepackage{hyperref}

\title{IGCV$3$: Interleaved Low-Rank Group Convolutions
	for Efficient Deep Neural Networks}

\addauthor{Ke Sun}{sunk@mail.ustc.edu.cn}{1}
\addauthor{Mingjie Li}{homles11@mail.ustc.edu.cn}{1}
\addauthor{Dong Liu}{dongeliu@ustc.edu.cn}{1}
\addauthor{Jingdong Wang}{jingdw@microsoft.com}{2}

\addinstitution{
	University of Science and Technology of China\\
	Anhui, China
}
\addinstitution{
	Microsoft Research Asia,\\
	Beijing, China
}

\runninghead{Student, Prof, Collaborator}{BMVC Author Guidelines}

\begin{document}
	
	\maketitle
	
	\begin{abstract}
		In this paper, we are interested in building lightweight and
		efficient convolutional neural networks.
		Inspired by the success of two design patterns,
		composition of structured sparse kernels, e.g., interleaved group convolutions (IGC),
		and composition of low-rank kernels, e.g., bottle-neck modules,
		we study the combination of such two design patterns, using the composition of structured sparse low-rank kernels,
		to form a convolutional kernel.
		Rather than introducing a complementary condition over channels,
		we introduce a loose complementary condition,
		which is formulated by imposing the complementary condition
		over super-channels,
		to guide the design for
		generating a dense convolutional kernel.
		The resulting network is called IGCV$3$.
		We empirically demonstrate that the combination of low-rank and sparse kernels boosts the performance
		and the superiority of our proposed approach
		to the state-of-the-arts, IGCV$2$ and MobileNetV$2$
		over image classification on CIFAR and ImageNet
		and object detection on COCO. Code and models are available at \url{https://github.com/homles11/IGCV3}. 
	\end{abstract}
	
	\section{Introduction}
	\label{sec:intro}
	
	There have been imperative demands for portable and efficient deep convolutional neural networks
	with high accuracies in vision applications.
	The recent efforts include two paths,
	network compression:
	approximate pre-trained models
	by pruning superfluous connections and channels
	or decomposing convolutional matrices;
	and lightweight network design:
	design less-redundant kernels for constructing networks
	trained from scratch. 
	
	We are interested
	in designing lightweight networks
	using less-redundant kernels to form a dense convolutional kernel.
	There are two main representative schemes,
	either of which has independently shown the success in building small networks.
	One scheme is to use a sequence of low-rank convolutional kernels to compose a linear kernel
	or nonlinear kernel with intermediate nonlinear activations included in the sequence of low-rank kernels,
	e.g., bottleneck~\cite{he2016deep} and inverted residual block~\cite{sandler2018inverted}.
	The other scheme is to use a sequence of sparse (and possible dense) kernels 
	to compose a kernel,
	e.g., interleaved group convolutions,
	MobileNetV1 and Xception,
	and deep roots~\cite{zhang2017interleaved,howard2017mobilenets,Chollet16a,IoannouRCC16}.
	
	In this paper,
	we design a modularized convolutional block,
	which simultaneously explores both low-rank and sparse kernels
	to compose a dense convolutional kernel.
	We start from IGCV$2$~\cite{ISSC18}, 
	which is composed of a channel-wise spatial convolution,
	group point-wise convolutions,
	and intermediate permutation operations.
	We introduce the low-rank design pattern
	into group point-wise convolutions,
	where one group convolution expands the feature dimension,
	and the other group convolution projects the feature back.
	We introduce a \emph{loose} complementary condition over channels for constructing \emph{dense} composed kernels,
	which is equivalent to the \emph{strict} complementary condition, presented in IGCV$2$ \cite{ISSC18},
	imposed over super-channels (a set of channels),
	so as to handle the case
	that the number of input and output channels are different
	and avoid over-wide convolutions.
	The resulting network is called IGCV$3$.
	We empirically demonstrate that the combination of low-rank and sparse kernels boosts the performance
	and the superiority of our proposed approach 
	to the state-of-the-arts, IGCV$2$ and MobileNetV$2$,
	over image classification on CIFAR and ImageNet
	and object detection on COCO.
	
	\section{Related Work}
	
	To obtain portable and efficient models, most of works devote great efforts to reduce the redundancy in the convolutional kernels, which mainly lie in two extents: high numerical precision and superfluous weights.

	\noindent\textbf{Low-precision kernels.}  
	Approaches in this field quantize the weights of CNNs from floating point into lower bit-depth representations. These methodologies reduce the redundancies in the numerical precision, such as binarization \cite{courbariaux2016binarized} constraining the weights to either $+1$ or $-1$, trinarization \cite{li2016ternary,zhou2016dorefa,zhu2016trained} making weights to be ternary-valued and quantization \cite{HanMD15,zhou2017incremental} a general form to convert weights into a low-precision version.
	
	\noindent\textbf{Sparse or Low-rank kernels.} In this field, the methods adopt one sparse or low-rank kernel as an approximation of the original dense or high-rank kernel.
	(i) $Sparse\ kernels:$ Many regularizations are designed to adding the sparse constraint on the convolutional kernels, such as $L1$ or $L2$ regularization \cite{HanMD15,HanPTD15} and structured sparsity regularizer \cite{LiKDSG16,WenWWCL16,alvarez2016learning}. Group convolution is a pre-defined structured-sparse matrix, which is widely used in the mobile models \cite{XieGDTH16,ZhaoWLTZ16}. (ii) $Low$-$rank\ kernels.$ Pruning redundant weights or channels from pre-trained models \cite{park2016faster,LiKDSG16,he2017channel,LuoWL17} is a main branch in network compression. What's more, many works \cite{SimonyanZ14a,SzegedyVISW16,denton2014exploiting} replace a large kernel with small kernels is to reduce the ranks in the spatial domain.
	
	\noindent\textbf{Composition from multiple sparse or low-rank kernels.} Design the composition of sparse or low-rank kernels to keep the original dense connections.
	(i) \emph{Multiplying the sparse kernels:} IGCV$1$ \cite{zhang2017interleaved} is the product of two structured-sparse matrices and IGCV$2$ \cite{ISSC18} decomposes the dense kernel into more sparse kernels. And the permutation operations \cite{zhang2017interleaved} are adopted to keep dense connectivity between input and output channels, which can increase the expressive capacity of networks \cite{sharir2017expressive}.
	Xception \cite{Chollet16a} is an extreme case of IGCV$1$: a pointwise convolution followed by a channel-wise convolution.
	(ii) $Composition\ from\ low$-$rank\ kernels.$ In a clear case, a $3\times3$ kernel can be decomposed into a $3\times1$ convolution followed by a $1\times3$ convolution \cite{IoannouRSCC15, JaderbergVZ14, MamaletG12}, which approximates the dense kernel along the spatial domain. Bottleneck \cite{he2016deep,iandola2016squeezenet,sandler2018inverted}, if neglecting the intermediate ReLUs, can be viewed as the low-rank approximation along the output channel domain.
	
	\section{Our Approach}
	In this section, we firstly review prior works, IGC and MobileNets, and then describe our proposed IGCV$3$. Finally, we analyze the loose complementary condition and give a discussion on the architecture of IGCV$3$ blocks.
	
	\begin{figure}[htb!]
		\centering
		\subfigure[IGCV$1$]{
			\begin{minipage}{4cm}
				\centering
				\includegraphics[scale=0.35]{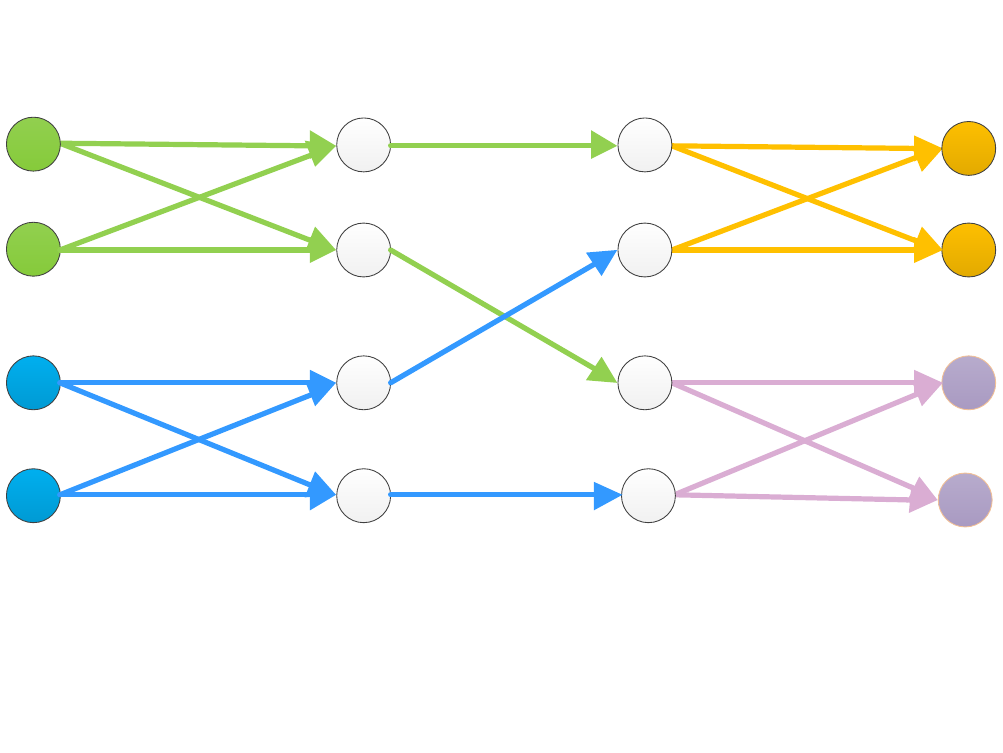}
				\label{fig:igc}
			\end{minipage}
		}
		\subfigure[Inverted Bottleneck]{
			\begin{minipage}{4cm}
				\centering
				\includegraphics[scale=0.35]{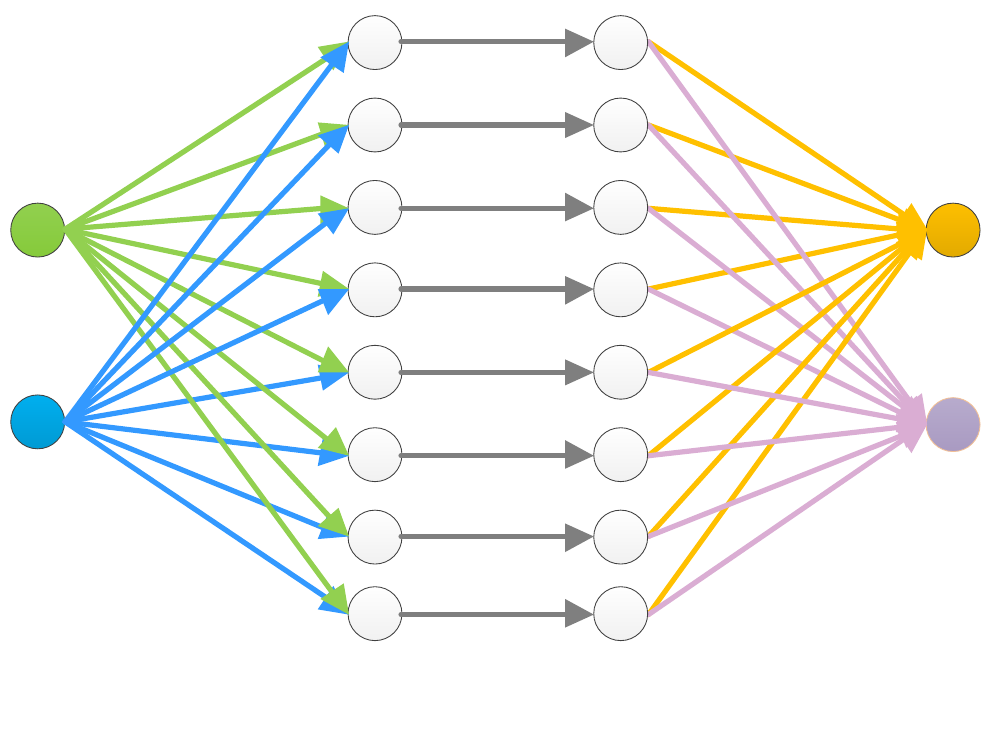}
				\label{fig:bottleneck}
			\end{minipage}
		}
		\subfigure[Inverted IGCV$3$]{
			\begin{minipage}{4cm}
				\centering
				\includegraphics[scale=0.35]{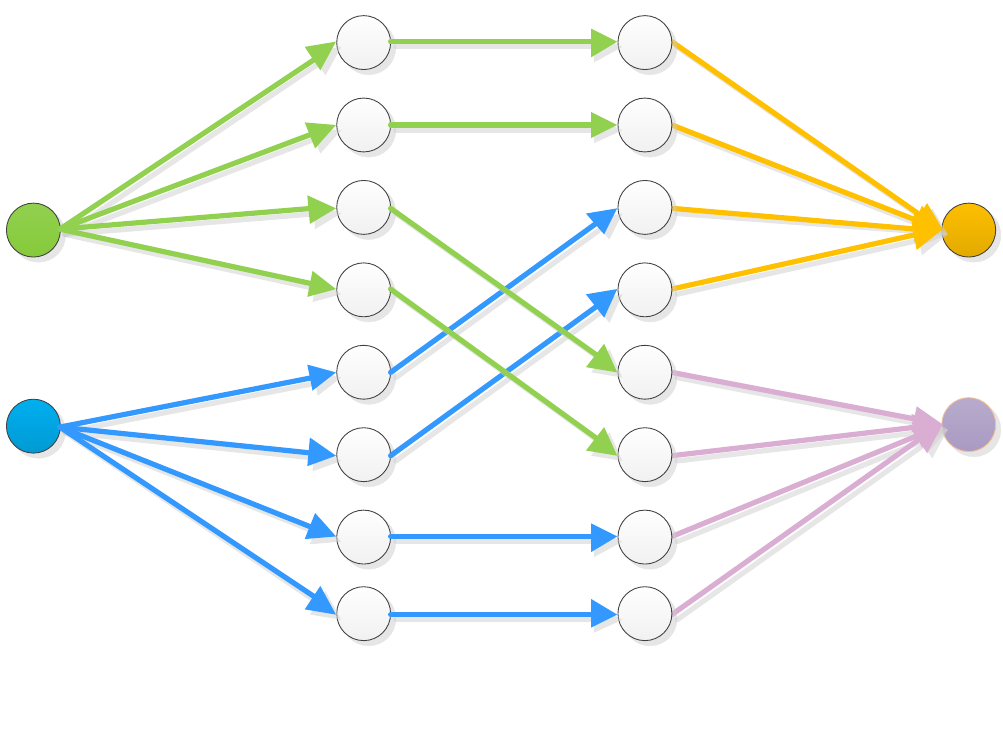}
				\label{fig:igcv3}
			\end{minipage}
		}
		\caption{The architectures of different blocks. Each circle denotes a channel and an arrow denotes a forward path. A set of arrows between two columns of circles can be viewed as a weight matrix. (a) IGCV$1$, the first and third matrices are sparse; (b) Inverted Bottleneck, the first and third matrices are low-rank; (c) Inverted IGCV$3$, the first and third matrices are sparse and low-rank. The second permutation in IGCV$1$ and IGCV$3$ is not shown here. Bottleneck and IGCV3 have the skip connection between the input and output.}
		\label{fig:diff_arch}
	\end{figure}
	\subsection{Prior Works}
	\textbf{Interleaved Group Convolution (IGCV$1$).}
	The IGCV$1$ block
	consists of primary and secondary group convolutions,
	which is mathematically formulated as follows:
	\begin{align}
	\mathbf{y} = \mathbf{P}^2\mathbf{W}^2 \mathbf{P}^1\mathbf{W}^1 \mathbf{x}.
	\label{eqn:IGC}
	\end{align}
	Here, the input response maps include $C$ channels and each channel has $K$ pixels. Let $\mathbf{x} \in \mathbb{R}^{(K\times C)\times 1}$ denote the corresponding input vector and $\mathbf{y} \in \mathbb{R}^{C\times1}$ denotes the output vector. The primary convolution is a group spatial convolution, whose kernel size is $K$, e.g., $K=9$ for $3\times 3$ kernels. $\mathbf{P}^1$ and $\mathbf{P}^2$ are permutation matrices~\cite{zhang2017interleaved}.
	The kernel matrices $\mathbf{W}^1$ and $\mathbf{W}^2$ are block-wise sparse,
	\begin{align}
	\mathbf{W}^i =  \begin{bmatrix}
	\mathbf{W}_{1}^i & \boldsymbol{0} &  \boldsymbol{0} &  \boldsymbol{0} \\[0.3em]
	\boldsymbol{0} & \mathbf{W}_{2}^i  & \boldsymbol{0} & \boldsymbol{0} \\[0.3em]
	\vdots & \vdots & \ddots  & \vdots \\[0.3em]
	\boldsymbol{0} & \boldsymbol{0} & \boldsymbol{0} & \mathbf{W}_{G_i}^i
	\end{bmatrix},
	\mathbf{x} =  \begin{bmatrix}
	\mathbf{x}^{1}\\[0.3em]
	\mathbf{x}^{2}  \\[0.3em]
	\vdots \\[0.3em]
	\mathbf{x}^{C}
	\end{bmatrix},
	\label{eqn:groupconvolution}
	\end{align}
	where $\mathbf{x}^{c}=[x_1^c,x_2^c,...,x_{K}^c]$ is a column vector. $\mathbf{W}_{g}^i$ ($i=1$ or $2$) is
	the kernel matrix
	over the corresponding channels
	in the $g$th branch,
	$G_i$ is the number of branches
	in the $i$th group convolution.
	In the case suggested in~\cite{zhang2017interleaved},
	the primary group convolution is a
	group $3\times 3$
	convolution with $G_1 = \frac{C}{2}$ groups,
	and $\mathbf{W}^1_g$ is a matrix of size $2 \times (2K)$.
	The secondary group convolution is a group $1\times 1$ convolution with $G_2 = 2$ groups,
	where $\mathbf{W}^2_1$ and $\mathbf{W}^2_2$
	are both dense matrices of size $\frac{C}{2} \times \frac{C}{2}$.

	\noindent\textbf{Interleaved Structured Sparse Convolution (IGCV$2$).}
	IGCV$2$ extends IGCV$1$
	by decomposing the convolution matrix into more structured sparse matrices:
	\begin{align}
	\mathbf{y} =~& \mathbf{P}_L\mathbf{W}_L \mathbf{P}_{L-1}\mathbf{W}_{L-1} \dots \mathbf{P}_1\mathbf{W}_1 \mathbf{x} \\
	=~& (\prod\nolimits_{l=L}^1 \mathbf{P}_l \mathbf{W}_l) \mathbf{x}.
	\label{eqn:ISSC}
	\end{align}
	Here $W_1$ corresponds to a channel-wise spatial convolution,
	and $W_l, l\in\{1,2,...,L\}$ corresponds to group point-wise convolutions.
	
	\noindent\textbf{MobileNetV$1$.}
	A MobileNetV$1$ block consists of a channel-wise spatial convolution
	and a point-wise convolution. The mathematical formulation
	is given as follows:
	\begin{align}
	\mathbf{y} = \mathbf{W}^2\mathbf{W}^1 \mathbf{x},
	\label{eqn:mnv1}
	\end{align}
	\noindent where $\mathbf{W^1}$ and $\mathbf{W^2}$ corresponds
	to the channel-wise and point-wise convolution respectively.
	It is an extreme case of IGCV$1$~\cite{zhang2017interleaved}:
	both channel-wise and point-wise convolutions
	are extreme group convolutions.
	
	\noindent\textbf{MobileNetV$2$.}
	A MobileNetV$2$ block consists of a dense pointwise convolution, a channel-wise spatial convolution, and a dense pointwise convolution.
	It uses an inverted bottleneck:
	the first pointwise convolution
	increases the width and
	the second one reduces the width.

	For convenience, we convert the input $\mathbf{x}$ to $\mathbf{\hat{x}}$, and accordingly the bottleneck is mathematically formulated:
	\begin{align}
	\mathbf{y} = \mathbf{W}^2\mathbf{W}^1\mathbf{W}^0 \mathbf{\hat{x}},
	\label{eqn:mnv2}
	\end{align}
	\begin{align}
	\mathbf{W}^0 =  \begin{bmatrix}
	\mathbf{\hat{w}}_{1,1}^\mathrm{T} & \boldsymbol{0} &  \boldsymbol{0} &  \boldsymbol{0}
	\\[0.3em]
	\boldsymbol{0} & \mathbf{\hat{w}}_{1,2}^\mathrm{T}  & \boldsymbol{0} & \boldsymbol{0} \\[0.3em]
	\vdots & \vdots & \ddots  & \vdots \\[0.3em]
	\boldsymbol{0} & \boldsymbol{0} & \boldsymbol{0} & \mathbf{\hat{w}}_{C_{int},K}^\mathrm{T}
	\end{bmatrix},
	\mathbf{\hat{x}} =  \begin{bmatrix}
	\mathbf{\hat{x}}_{1,1}\\[0.3em]
	\mathbf{\hat{x}}_{1,2}  \\[0.3em]
	\vdots \\[0.3em]
	\mathbf{\hat{x}}_{C_{int},K}
	\end{bmatrix},
	\label{eqn:mmnv2}
	\end{align}
	where $\mathbf{W}^1 \in \mathbb{R}^{C_{int}\times (K\times C_{int})}$ corresponds to the channel-wise $3 \times 3$ convolution, the kernel including $K=9$ spatial positions, and $\mathbf{W}^0 \in \mathbb{R}^{(K\times C_{int})\times (K\times C_{int}\times C)}$ and $\mathbf{W}^2 \in \mathbb{R}^{C\times C_{int}}$ are two low-rank matrices, which correspond to the two dense point-wise convolutions. $C_{int}$ represents the intermediate width. $\mathbf{\hat{x}}\in \mathbb{R}^{(K\times C_{int} \times C)\times 1}$ is a column vector and $\mathbf{\hat{x}}_{j,k} = [x^1_k,x^2_k,...,x^{C}_k]$, $j \in \{1,2,...,C_{int}\}$ is the $k$th spatial position across all the input channels for the $j$th output channel. $\mathbf{\hat{w}}_{j,k}=[w_j^1,w_j^2,...,w_j^{C}], k\in\{1,2,...,K\}$ represents the $j$th kernel of the first convolution, which is shared by $K$ spatial positions.
	
	\begin{figure}[t]
		\centering
		\includegraphics[scale=0.5]{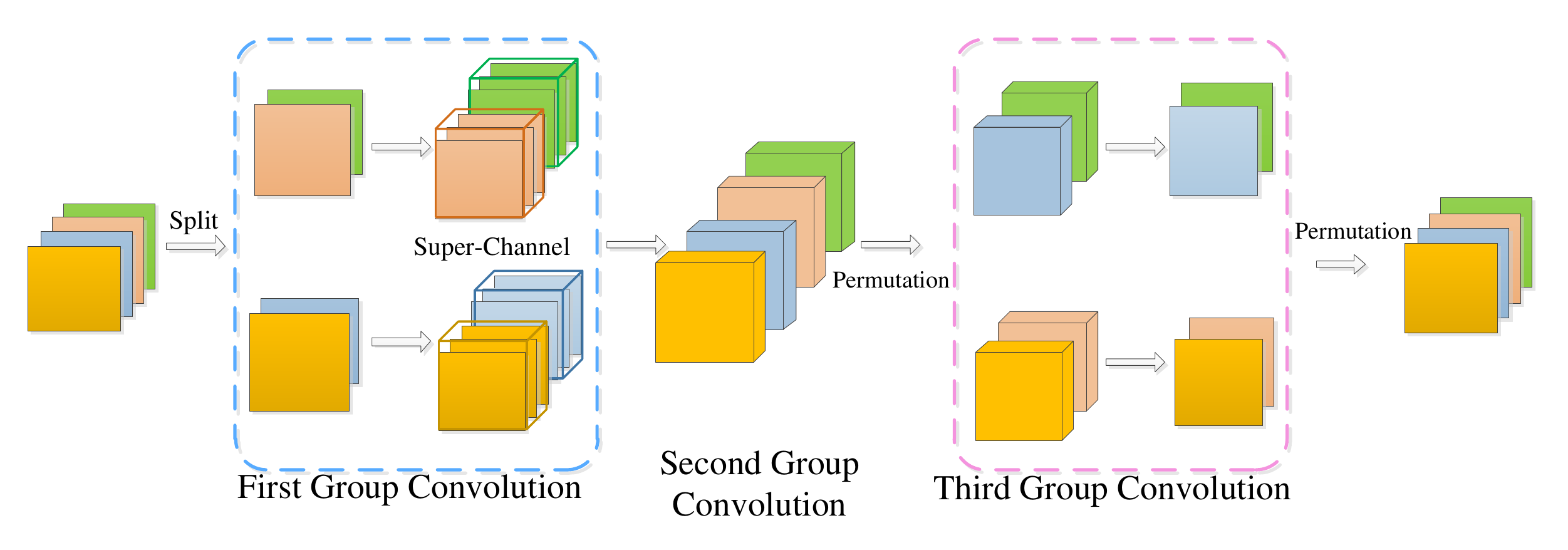}
		\caption{Illustrating the interleaved branches in IGCV$3$ block. The first group convolution is a group $1\times1$ convolution with $G_1=2$ groups. The second is a channel-wise spatial convolution. The third is a group $1\times1$ convolution with $G_2=2$ groups.}
		\label{fig:igc-v3-path}
	\end{figure}
	\subsection{Interleaved Low-Rank Group Convolutions}
	The proposed Interleaved Low-Rank Group Convolutions,
	named IGCV$3$, extend IGCV$2$
	by using low-rank group convolutions
	to replace group convolutions in IGCV$2$.
	It consists of a channel-wise spatial convolution,
	a low-rank group point-wise convolution with $G_1$ groups
	that reduces the width
	and a low-rank group point-wise convolution with $G_2$ groups
	which expands the width back.
	The mathematical formulation is given as follows,
	\begin{align}
	\mathbf{y} = \mathbf{P}^2\mathbf{W}^2\mathbf{P}^1\mathbf{W}^1\mathbf{\hat{W}}^0 \mathbf{\hat{x}}.
	\label{eqn:igcv3}
	\end{align}
	Here, $\mathbf{P}^1$ and $\mathbf{P}^2$ are permutation matrices
	similar to permutation matrices given in~\cite{zhang2017interleaved}.
	$\mathbf{W}^1$ corresponds to the channel-wise $3\times 3$ convolution.
	$\mathbf{\hat{W}}^0$ and $\mathbf{W}^2$ are low-rank structured sparse matrices. The two low-rank sparse matrices are mathematically formulated as follows,

	\begin{align}
	\mathbf{\hat{W}^0} =  \begin{bmatrix}
	\mathbf{\hat{w}}_{1,1}^{1} & \boldsymbol{0} &  \boldsymbol{0} &  \boldsymbol{0}
	\\[0.3em]
	\boldsymbol{0} & \mathbf{\hat{w}}_{1,2}^{1}  & \boldsymbol{0} & \boldsymbol{0} \\[0.3em]
	\vdots  & \vdots & \ddots  & \vdots \\[0.3em]
	\boldsymbol{0} & \boldsymbol{0} & \boldsymbol{0}& \mathbf{\hat{w}}_{C_{int},K}^{G_1}
	\end{bmatrix},
	\mathbf{{W}^2} =  \begin{bmatrix}
	\mathbf{{W}}_{1}^{2} & \boldsymbol{0} &  \boldsymbol{0} &  \boldsymbol{0}
	\\[0.3em]
	\boldsymbol{0} & \mathbf{{W}}_{2}^{2}  & \boldsymbol{0} & \boldsymbol{0} \\[0.3em]
	\vdots & \vdots & \ddots  & \vdots \\[0.3em]
	\boldsymbol{0} & \boldsymbol{0} & \boldsymbol{0} & \mathbf{{W}}_{G_2}^{2}
	\end{bmatrix}.
	\label{eqn:migcv3}
	\end{align}
	Here, $\mathbf{\hat{w}}^g_{j,k} \in \mathbb{R}^{C}$, a row vector including $\frac{C}{G_1}$ non-zero weights, corresponds to the kernels for the $g$th branch in the first group convolution, e.g. $\mathbf{\hat{w}}^1_{j,k} = [w_j^1,w_j^2,...,w_j^{\frac{C}{G_1}},0,...,0]^\mathrm{T}$, which increases the number of channels. $\mathbf{{W}}_{g}^{2}$ corresponds to the $g$th branch in the second group convolution, which reduces the width back to the initial width.
	
	\noindent\textbf{Construct a dense composed kernel.}
	Similar to IGCV$2$, we also aim to design the two group convolutions such that
	the composed pointwise convolutional kernel, $\mathbf{P}^2\mathbf{W}^2\mathbf{P}^1\mathbf{W}^1$, is dense. Different from IGCV$2$, the number of input and output channels in the low-rank group convolution are not the same, so the permutation operation cannot work as before.
	
	We divide input, output and intermediate channels into a set of (e.g., $C_s$) super-channels, where the super-channel for input and output response maps contains $\frac{C}{C_s}$ channels and for intermediate response maps the super-channel contains $\frac{C_{int}}{C_s}$ channels.
	The sketch is shown in Fig \ref{fig:igc-v3-path}, where super-channels are represented as 3-D boxes in different colors. 
	Then, we review the \emph{strict} complementary condition proposed in \cite{ISSC18} and describe the \emph{loose} complementary condition based on the concept of super-channel.
	
	\newtheorem{condition}{Condition}
	
	\begin{condition}[Strict complementary condition]
		The two group convolutions are thought complementary
		if
		the channels in the input and output response maps and 
		in the intermediate response maps  
		lying in the same branch
		in one group convolution
		lie in different branches
		and come from all the branches in the other group convolution.
	\end{condition}
	
	\noindent\textbf{Loose complimentary condition.} 
	We empirically observed that 
	over-sparse convolutions lead to over-wide response maps 
	and thus too large memory cost,
	but do not lead to higher accuracy,
	which is consistent to the empirical results in IGCV$1$~\cite{zhang2017interleaved} and IGCV$2$~\cite{ISSC18}.
	Therefore,
	we relax the strict complementary condition 
	to a loose version.
	\begin{condition}[Loose complementary condition]
		The two group convolutions are thought complementary
		if
		the super-channels lying in the same branch
		in one group convolution
		lie in different branches
		and come from all the branches in the other group convolution.
	\end{condition}
	
	\subsection{Discussions and Analysis}
	\noindent\textbf{Linear, nonlinear, and inverted IGCV$3$.} 
	It is shown in IGCV$1$~\cite{zhang2017interleaved}
	that the linear version,
	i.e., no intermediate ReLU is included in the block,
	performs better than the very deep networks
	which include lots of ReLU activations.
	Similar observations are also obtained
	in the bottleneck block~\cite{he2016deep} and Xception~\cite{Chollet16a}.
	On the other hand, for not very deep networks,
	e.g., there are only $10+$ IGCV$3$ blocks,
	the non-linear version, i.e., 
	there are one or more intermediate ReLUs in the block,  
	performs better. 
	We provide the ablation study in Sec. \ref{sec:ab-study}. 
	
	In our experiments, we do not see difference between the normal version and the inverted version of bottleneck. We adopt the inverted IGCV$3$ block to save the memory footprints during the training and inference process (shown in Fig~\ref{fig:igcv3}):
	low-rank group convolution (width increasing),
	channel-wise spatial convolution,
	and low-rank group convolution (width reduction)
	and accordingly the identity connection is built between two low-dimensional representations, which is similar to inverted bottleneck presented in MobileNetV$2$ \cite{sandler2018inverted}.
	
	\noindent\textbf{Wider and deeper IGCV$3$ networks.} 
	We stack our IGCV$3$ blocks to form a deep network and provide two versions for fair comparisons with MobileNetV$2$.
	One is to let the depth (the number of blocks) be the same and to 
	widen the network. In our implementation, 
	$C_s=4$ super-channels is used to form group convolutions ($G_1=4$ and $G_2=4$).
	The other one is to let the width (of the corresponding block) be the same 
	and to deepen the network. 
	In our implementation, $C_s = 2$ super-channels is used ($G_1=2$ and $G_2=2$).
	The empirical study is given in Sec. \ref{sec:ab-study}.
	
	\section{Experiments}
	\label{sec:exp}
	\subsection{Datasets}
	\noindent\textbf{CIFAR.}
	The CIFAR datasets, CIFAR-$10$~\cite{Cifar10} and CIFAR-$100$~\cite{Cifar100},
	are subsets of the $80$ million tiny images.
	Both datasets contain $60000$ $32\times32$ color images with $50000$ images for training and $10000$ images for test. 
	The standard data augmentation scheme we adopt is widely used for these datasets~\cite{HeZRS16, LeeXGZT15, HuangLW16a, LarssonMS16a, LinCY13, RomeroBKCGB14, SpringenbergDBR14, SrivastavaGS15}: we zero-pad the images with $4$ pixels on each side, and then randomly crop them to produce $32\times32$ images, followed by horizontally mirroring half of the images and normalizing them by using the channel means.
	
	For training, we use the SGD algorithm and train all networks from scratch. We initialize the weights similar to \cite{HeZRS16, HeZRS16ECCV}, and set the weight decay as 0.0001 and the momentum as 0.9. The initial learning rate is 0.1 and is reduced by a factor 10 at the 200, 300 and 350 training epochs. Each training batch consists of 64 images on four asynchronous GPUs.
	
	\noindent\textbf{ImageNet.} The ILSVRC 2012 classification dataset~\cite{deng2009imagenet} contains over 1.2 million training images and 50,000 validation images, and each image is labeled from 1000 categories. We adopt the same data augmentation scheme as in \cite{HeZRS16,HeZRS16ECCV}. We use SGD to train the networks with the same hyperparameters (batch size $=96$, weight decay $=0.00004$ and momentum $=0.9$) on 4 GPUs. We train the models for $480$ epochs with extra $50$ epochs for retraining. We start from a learning rate of $0.045$, and then scale it by $0.98$ every epochs.
	
	\noindent\textbf{MSCOCO.} We evaluate and compare the performance of IGCV3 and MobileNetV2 for object detection on COCO dataset~\cite{lin2014microsoft}, which are used as a backbone for SSDLite and SSDLite2 respectively. As in previous work~\cite{sandler2018inverted}, we train using the union of 80k training images and a 35k subset of val images (trainval35k) and evaluate on test-dev.
	We train the models for $240$ epochs on MXNet. We start from a learning rate of $0.004$, and then divide it by 10 every 60 epochs. The input resolution is shown in Table \ref{det:coco}.
	\subsection{Comparisons with IGCV$1$ and IGCV$2$.}
	We compare IGCV$3$ with IGCV$1$ and IGCV$2$ on CIFAR and ImageNet. The models of IGCV$1$ and IGCV$2$ replace the blocks in MobileNetV$1$ with their proposed blocks. Please check the details in \cite{zhang2017interleaved,ISSC18}. IGCV$3$ adopts two group convolutions with $G_1=2$ and $G_2=2$ respectively for the deeper version (IGCV$3$-D). 
	
	\begin{table}[htb!]
		\begin{center}
			\begin{tabular}{|l||c|c|c|c|}
				\hline
				& \#Params& CIFAR-10 & CIFAR-100& ImageNet\\
				\hline\hline
				IGCV$1$ (our impl.)    & 2.2M  &   $91.77\pm0.14$    & $70.07\pm0.17$& -\\
				IGCV$2$ (our impl.) & 2.2M  & $94.76\pm0.11$ &$77.45\pm0.35$& 70.7  \\
				IGCV$3$-D & 2.2M & $\mathbf{94.96\pm{0.07}}$ &$\mathbf{77.95{\pm0.39}}$&$\mathbf{72.1}$ \\
				\hline
			\end{tabular}
		\end{center}
		\caption{Comparisons of different IGC blocks for classification accuracy on CIFAR and ImageNet. All the models keep the same number of parameters. Experiments are repeated five times, and the results are shown as $mean\pm std$. The parameter numbers are calculated without counting the final FC-layer. The results of IGCV$1$ and IGCV$2$ on CIFAR are reproduced by ourselves.}
		\label{tab:diiff_igc}
	\end{table}
	
	The comparisons of IGC blocks are shown in Table \ref{tab:diiff_igc}. IGCV$3$ outperforms the prior works slightly on CIFAR datasets, and achieves significant improvement about 1.5\% on ImageNet. 
	By introducing low-rank design into the sparse kernels, IGCV$3$ reduces the redundancies in feature dimensions and increases the depth of network to improve the generalization and the overall performance.
	\subsection{Comparisons with Other Mobile Networks.}
	We compare IGCV$3$ with MobileNetV$2$ and other mobile networks for classification and detection on several benchmarks. IGCV$3$ adopts two group convolutions with $G_1=2$ and $G_2=2$ for the deeper version (IGCV$3$-D) and with $G_1=4$ and $G_2=4$ for the wider version (IGCV$3$-W). Due to the limitation, we reproduce MobileNetV$2$ on 4 GPUs not proposed 16 GPUs. We try our best to train MobileNetV$2$ and provide the reproduced results as (our impl.). We adopt the settings of the best reproduced MobileNetV$2$ to train our models.
	\begin{table}[htb!]
		\centering
		\begin{tabular}{|l||c|c|c|}
			\hline
			& \#Params& CIFAR-10 & CIFAR-100\\
			\hline\hline
			MobileNetV$2$ (our impl.) & 2.1M & $94.56\pm{0.16}$ & $77.09\pm{0.17}$\\
			IGCV$3$-D $0.5\times$ & 1.1M & $94.73\pm0.14$  & $77.29\pm0.25$ \\
			IGCV$3$-D $0.7\times$ & 1.5M & $94.92\pm0.16$ & $77.83\pm0.38$ \\
			IGCV$3$-D $1.0\times$ & 2.2M & $\mathbf{94.96\pm{0.07}}$ &$\mathbf{77.95\pm0.39}$ \\
			\hline
		\end{tabular}
		\caption{Classification accuracy comparisons of MobileNetV$2$ and IGCV$3$ on CIFAR datasets. "Network s$\times$" means reducing the number of parameters in "Network $1.0\times$" by s times. For IGCV$3$-D, we reduce the number of blocks to control the count of parameters. The parameter numbers are calculated without counting the final FC-layer.}
		\label{tab:diff_mn_igcv3}
	\end{table}
	
	\noindent\textbf{CIFAR classification.} The classification accuracy comparisons of MobileNetV$2$ and IGCV$3$ are presented in Table \ref{tab:diff_mn_igcv3}. Introducing sparse design into low-rank kernels, IGC V3 further reduces the superfluous weights in convolutional kernels, and also makes the best use of parameters to improve the performance. IGCV$3$ outperforms MobileNetV$2$ a lot with the similar number of parameters. Moreover, IGCV$3$ with 50\% parameters still achieves a better performance, which has the same depth as MobileNetV2. The reason may be that the number of ReLU is half of MobileNet V2.
	
	\begin{table}[h]
		\centering
		\begin{tabular}{|l||c|c|c|}
			\hline
			& \#Params& MAdds&ImageNet\\
			\hline\hline
			MobileNetV$2$ (0.7)(our impl.)&2.8M& 210M &66.51\\
			IGCV$3$-D (0.7) & 2.8M & 210M & \textbf{68.45} \\
			\hline\hline
			MobileNetV$1$ \cite{howard2017mobilenets}&4.2M&575M&70.6\\
			ShuffleNet (1.5) \cite{zhang2017shufflenet} &3.4M&292M&71.5\\
			MobileNetV$2$ \cite{sandler2018inverted}&3.4M&300M&72.0\\
			MobileNetV$2$ (our impl.) &3.4M&300M&71.3\\
			IGCV3-D & 3.5M & 318M & \textbf{72.2} \\
			\hline\hline
			ShuffleNet (2.0) \cite{zhang2017shufflenet}&5.4M&524M&73.7\\
			NasNet-A \cite{ZophVSL17}&5.3M&564M&74.0\\
			MobileNetV$2$ (1.4) \cite{sandler2018inverted}&6.9M&585M&\textbf{74.7}\\
			MobileNetV$2$ (1.4)(our impl.)&6.9M&585M&73.76\\
			IGCV$3$-D (1.4) & 7.2M  & 610M&74.55 \\		
			\hline
		\end{tabular}
		\caption{Comparisons of different mobile networks for classification on ImageNet. We count the total number of Multiply-Adds for ops. "Network ($\alpha$)" means scaling the
			number of filters in "Network (1.0)" by
			$\alpha$ times thus overall
			complexity will be roughly $\alpha^2$
			times of "Network ($\alpha$)".}
		\label{tab:diff_imagenet}
	\end{table}
	\noindent\textbf{ImageNet classification.} We compare IGCV$3$ with other mobile models on ImageNet, shown in Table \ref{tab:diff_imagenet}. With the similar computation complexity, IGCV$3$ always outperforms other networks. IGCV$3$-D (1.4) achieves a superior performance under the same training settings (74.55\% vs 73.76\%), though marginally worse than the original MobileNetV$2$ (1.4). Due to the difference of the final FC-layer, IGCV$3$-D has the different number of parameters in Table \ref{tab:diff_imagenet} from Table \ref{tab:diiff_igc} and \ref{tab:diff_mn_igcv3} ($1.3$M (1000 classes), $13$k (100 classes) and $1.3$k (10 classes)).
	
	\begin{table}[ht]
		\centering
		\begin{tabular}{|l||c|c|c|}
			\hline
			& Input Size & \#Params& mAP\\
			\hline\hline
			SSD-300 \cite{liu2016ssd}                          &300x300&36.1M&23.2\\
			SSD-512 \cite{liu2016ssd}                          &512x512&36.1M&26.8\\
			YOLOV$2$ \cite{redmon2017yolo9000}                           &-&50.7M&21.6\\
			MobileNetV$1$ SSDLite \cite{howard2017mobilenets}                 &320x320&5.1M&22.2\\
			MobileNetV$2$ SSDLite \cite{sandler2018inverted}                 &320x320&4.3M&22.1\\
			\hline\hline
			MobileNetV$2$ SSDLite (our impl.)&320x320& 4.3M & $22.1$ \\
			IGCV$3$-W SSDLite2       &320x320& 4.0M & $22.2$ \\
			\hline
		\end{tabular}
		\caption{Comparisons of different methods for detection on COCO. The models are evaluated with the different input size. Reproduced MobilenetV2 and IGCV3 are trained and evaluated with the input resolution of $320\times320$.}
		\label{det:coco}
	\end{table}
	\noindent\textbf{Detection on COCO.} We extend IGCV$3$ to be a backbone for detection networks. SSDLite is proposed in \cite{sandler2018inverted}, we follow the original framework, but replace all the feature extraction blocks with IGCV3, denoted by "SSDLite2".
	The comparisons of different methods for detection are shown in Table \ref{det:coco}. We adopt IGCV$3$-W as our backbone for extracting the features from the layers at the same depth as MobileNetV2. 
	IGCV3 is slightly better than MobileNetV$2$ with fewer parameters, and outperforms YOLOV$2$ 0.6\% mAP with much fewer number of parameters.
	
	\section{Ablation Study}
	\label{sec:ab-study}
	We explore three main design choices: (i) deeper and wider networks; 
	(ii) intermediate ReLUs; (iii) \#branches in group convolutions. These three elements determine the architecture of the IGCV3 block and the whole network.
	
	\noindent\textbf{Deeper and wider networks.}\label{deep_wider} Most of related works enlarge the width of networks, which output high-dimensional features capturing more information. 
	\begin{table}[htb!]
		\centering
		\begin{tabular}{|l||c|c|c|c|}
			\hline
			&\#Params& CIFAR-10& CIFAR-100 &ImageNet\\
			\hline\hline
			IGCV$3$-W & 2.1M   & $94.68\pm0.21$   & $76.48\pm0.25$     & 71.1 \\
			IGCV$3$-Widest & 2.1M   & $94.62\pm0.19$   & $76.39\pm0.27$     & - \\
			IGCV$3$-D & 2.2M     & $\mathbf{94.88\pm0.08}$   &   $\mathbf{77.95\pm0.39}$   &$\mathbf{72.2}$\\
			\hline
		\end{tabular}
		\caption{Comparisons of deeper and wider networks on CIFAR and ImageNet. IGCV$3$-Widest is designed by following the strict complementation condition.} 
		\label{wd_dp}
	\end{table}
	
	Comparisons between deeper and wider networks are shown in Table \ref{wd_dp}. It can be observed that the deeper version significantly outperforms the wider version, which is as expected: (i) there are redundancies in feature dimensions, so further enlarging the width cannot bring about gains; (ii) the networks built by stacking bottlenecks improve the final performance with the increasing of depth, consistent to the observations in \cite{HeZRS16, HuangLW16a}.
	\begin{table}[htb!]
		\centering
		\begin{tabular}{|c||c|c|c|}
			\hline
			& \#Params& CIFAR-10 & CIFAR-100\\
			\hline\hline
			G1x1\_ReLU\_Cw3x3\_ReLU\_G1x1 & 2.2M  & $94.44\pm0.11$&$76.61\pm0.29$  \\
			G1x1\_Cw3x3\_ReLU\_G1x1 & 2.2M  & $\mathbf{94.88\pm0.07}$&$\mathbf{77.95\pm0.39}$  \\
			G1x1\_Cw3x3\_G1x1\_ReLU & 2.2M  & $93.76\pm0.08$&$75.53\pm0.25$  \\
			\hline
		\end{tabular}
		\caption{Comparisons of different IGCV3 blocks for classification on CIFAR. The first column represents the architectures, $G1\times1$ and $Cw$ denote the group $1\times1$ convolution and the channel-wise convolution respectively. The positions of ReLUs in the first block is the same as the MobileNetV2 block and the second is adopted in our IGCV3 block.
			Each convolution is followed by a BN.}
		\label{non-linear}
	\end{table}
	
	\noindent\textbf{Intermediate ReLUs.} We explore the positions of non-linearities and the results are shown in Table \ref{non-linear}. The second block (IGCV$3$ block) has obvious advantages over other blocks. The first block includes two ReLUs, and thus the resulting network has two times ReLUs more than others, which may be a reason for its worse performance. The third block is equivalent to a convolutional kernel followed by a ReLU, which is worse than IGC blocks. 
	
	\noindent\textbf{\#Branches in group convolutions.} It's easy to find that there are many possible combinations of $G_1$ and $G_2$ for two group convolutions satisfying the loose complementary condition, only if the product of $G_1$ and $G_2$ is \emph{the common divisor} of the number of input and output channels. Table \ref{sparsity} provides another two cases.
	\begin{table}[htb!]
		\centering
		\begin{tabular}{|c||c|c|c|}
			\hline
			& \#Params& CIFAR-10 & CIFAR-100\\
			\hline\hline
			IGCV3-D($G_1=2$,$G_2=2$) & 2.2M  & $94.88\pm0.07$ & $77.95\pm0.39$ \\
			IGCV3-D($G_1=2$,$G_2=4$) & 2.2M  & $\mathbf{94.89\pm0.11}$     &$\mathbf{78.12\pm0.36}$  \\
			IGCV3-D($G_1=4$,$G_2=2$) & 2.2M  & $94.71\pm0.26$     &$77.84\pm0.40$  \\
			\hline
		\end{tabular}
		\caption{Comparisons of IGCV$3$ blocks with different branches for classification on CIFAR.}
		\label{sparsity}
	\end{table}
	
	From the results, shown in Table \ref{sparsity}, we can find that the first group convolution prefers to be denser. The third group convolution projects the high-dimensional features back to the low-dimensional space, which results in information loss. Therefore, reducing more kernels may have little effect on its performance. In these cases, the IGCV$3$ blocks in the resulting networks share the same settings, e.g. $G_1=2$, $G_2=2$. Actually, IGCV$3$ blocks can adopt the different settings independently, which leads to a deeper network and achieves a better performance. In our experiments, we adopt $G_1=2$, $G_2=2$ to reduce the memory cost, which also achieves a good performance.
	
	\section{Conclusion}
	In this paper,
	we propose a novel block, interleaved low-rank group convolutions, 
	which enjoys the benefits of both low-rank and sparse convolutions. 
	We introduce super-channels and the loose complementary condition
	to guide the design. 
	Empirical results demonstrate the efficient of our models and the superiority over the state-of-the-arts, MobileNetV$2$ and IGCV$2$, on several benchmarks.
	
	\bibliography{egbib}
\end{document}